\documentclass[a4paper,conference]{IEEEtran}
\usepackage{cite}

\ifCLASSINFOpdf
  \usepackage[pdftex]{graphicx}
\else
\fi

\usepackage{amsmath}
\usepackage{fixltx2e}
\usepackage{stfloats}
\usepackage{url}
\usepackage{newfloat}
\usepackage{listings}
\usepackage{amsmath}
\usepackage{amssymb}
\usepackage{multirow}
\usepackage{xcolor}

\newcommand{\minisection}[1]{\vspace{0.04in} \noindent {\bf #1}\ \ }

\begin{document}
\title{Visual Transformers with Primal Object Queries for Multi-Label Image Classification}

\author{\IEEEauthorblockN{Vacit Oguz Yazici}
\IEEEauthorblockA{Computer Vision Center/Wide Eyes Technologies\\
Barcelona, Spain\\
Email: voyazici@cvc.uab.es}
\and
\IEEEauthorblockN{Joost van de Weijer}
\IEEEauthorblockA{Computer Vision Center,\\ Universitat Autonoma de Barcelona\\
Email: joost@cvc.uab.es}
\and
\IEEEauthorblockN{Longlong Yu}
\IEEEauthorblockA{Wide-Eyes Technologies\\Barcelona, Spain\\
Email: longyu@wide-eyes.it}}

\maketitle

\begin{abstract}
Multi-label image classification is about predicting a set of class labels that can be considered as orderless sequential data. Transformers process the sequential data as a whole, therefore they are inherently good at set prediction. The first vision-based transformer model, which was proposed for the object detection task introduced the concept of \emph{object queries}. Object queries are learnable positional encodings that are used by attention modules in decoder layers to decode the object classes or bounding boxes using the region of interests in an image. However, inputting the same set of object queries to different decoder layers hinders the training: it results in lower performance and delays  convergence. In this paper, we propose the usage of \emph{primal object queries} that are only provided at the start of the transformer decoder stack. In addition, we improve the mixup technique proposed for multi-label classification. The proposed transformer model with \emph{primal object queries} improves the state-of-the-art class wise F1 metric by 2.1\% and 1.8\%; and speeds up the convergence by 79.0\% and 38.6\% on MS-COCO and NUS-WIDE datasets respectively.
\end{abstract}

\IEEEpeerreviewmaketitle

\section{Introduction}
\label{sec:intro}
The task of predicting the presence of visual concepts in images is known as multi-label classification. The visual concepts in general refer to a set of objects, but could also refer to other visual concepts such as attributes. Multi-label classification is difficult because of the wide range of classes that is typically considered, the wide variety of scales in which these classes can occur, and the complex inter-dependencies between classes~\cite{bi2013efficient, wehrmann2018hierarchical}.

The field of multi-label image classification has seen much progress in recent years. Earlier works exploited graphical models to model label relations explicitly~\cite{ghamrawi2005collective, guo2011multi}. Then, CNN-RNN models were proposed~\cite{wang2016cnn, liu2017semantic, yazici2020orderless} to capture label correlations. Later, to learn label dependencies explicitly, graph convolutional networks were proposed~\cite{guo2019visual}. Even though significant progress has been made in multi-label image classification in recent years, systems still suffer from problems common to recursive methods (such as modeling long-term dependencies) or fail to capture the complex relations between the many visual concepts involved in multi-label classification. 

Transformers were first proposed in~\cite{vaswani2017attention}. Unlike recurrent models, transformers process data simultaneously. In case of recurrent models, if the decoding process is interrupted the decoder is forced to output a \emph{termination} token, which will cause the not-yet attended classes to be missed. Due to the non-sequential nature of transformers, they do not suffer from this problem. Therefore, although they were firstly used for various NLP tasks~\cite{devlin-etal-2019-bert, cornia2020meshed, zhu2018captioning}, they became also widely used for other tasks. The first visual CNN-transformer model for a vision task was proposed for object detection~\cite{carion2020end}. Recently, a novel and pure transformer model that uses a sequence of image patches as input data was proposed for image classification~\cite{dosovitskiy2021an}. Lanchantin et al.~\cite{lanchantin2021general} proposed the first transformer model for multi-label image classification.

\begin{figure}[!tb]
\centering
  \includegraphics[width=0.95\linewidth]{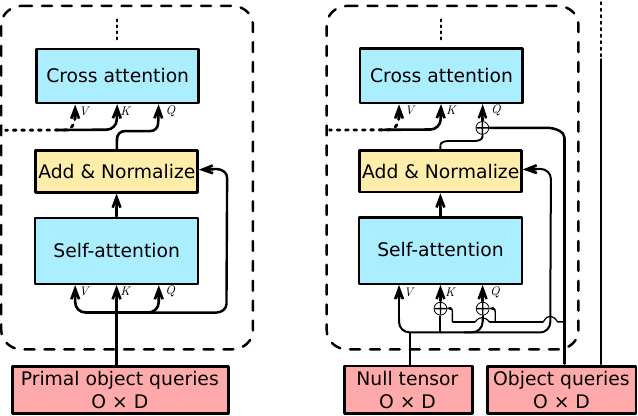}
  \caption{\small Comparison between the primal object queries and object queries. Primal object queries are given directly to the first decoder layer instead of being added as positional encodings in every decoder layer. This leads to faster convergence and better performance.}
  \label{fig:our_detr_model}
\end{figure}

Carion et al.~\cite{carion2020end} introduced the concept of \emph{object queries}. As a set of learnable positional encodings, they are added to query and key tensors in attention modules of every decoder layer. However, using the same set of object queries in different decoder layers is not optimal for training due to the different set of relations learned by each of the layers. As we empirically demonstrate in the experimental section, it leads to lower performance and slower convergence. Therefore, in this paper, we introduce \emph{primal object queries} that differ from standard object queries in the way that they are input to a transformer decoder stack (see Figure~\ref{fig:our_detr_model}). Moreover, we improve the mixup technique for multi-label classification to achieve significantly better results.

The main contributions in this paper are:
\begin{itemize}
    \item We introduce \emph{primal object queries} that obtain better results and yield faster convergence than standard object queries by 79.0\% and 38.6\% on MS-COCO and NUS-WIDE datasets respectively.
    \item We improve the mixup for multi-label classification and show that it results in a significant improvement.
    \item Evaluation shows that we improve the state-of-the-art class wise F1 score by 2.1\% and 1.8\% on the MS-COCO and NUS-WIDE datasets respectively.
\end{itemize}

\section{Related Work}
\label{sec:related_work}

\minisection{Transformers.}
Transformers were first proposed in~\cite{vaswani2017attention}, and since became the state-of-the-art approach for sequence-to-sequence tasks such as machine translation and visual question answering. The transformer self-attention module attends all the sequential data at once, hence handling the long sequences better than RNNs, which struggle with long-term dependencies. A challenging aspect of transformers is the large number of parameters they require, which require large amounts of data to fit properly. Since such large-scale training datasets are scarce, and an important limitation to use transformer models in many practical applications, Devlin et al.~\cite{devlin-etal-2019-bert} proposed a way of training transformers in an unsupervised manner on readily available large unsupervised text corpus, and showed that with a simple fine-tuning procedure state-of-the-art results could be achieved on various tasks. This led to a wider popularization of transformer models, and usage in areas different than the originally proposed ones, such as image captioning~\cite{cornia2020meshed, zhu2018captioning} and object detection~\cite{carion2020end}. Carion et al.~\cite{carion2020end} introduced the new concept of object queries to be given input for the transformer through the training. These object queries are learned through the training, and according to the analysis done by the authors, each object query learns to focus on different areas of images and box sizes. Dosovitskiy et al.~\cite{dosovitskiy2021an} proposed the first pure transformer model for image classification. They split an image into smaller patches and convert it to a sequential data (which also consists of tokens) to be processed by transformer encoder layers. Although they showed that transformers can give promising results for pure vision tasks, they still fell behind other convolutional models. Yuan et al.~\cite{yuan2021tokens} proposed a new tokenization system that reduced the number of parameters and achieved comparable results with other convolutional models. Recently, Lanchantin et al.~\cite{lanchantin2021general} introduced the first transformer model for multi-label image classification. They exploited self-attention modules to learn label dependencies for the purpose of predicting a set of labels given a set of masked label embeddings and image features.

\minisection{Multi-label Classification.}
Multi-label classification seeks to predict a variable number of labels for every single image, ideally capturing all the relevant visual concepts, such as objects or attributes, that appear on the image. Traditional methods for multi-label classification ignore label correlation, which can help boost the performance of certain under-represented classes. The few early works that tried to leverage label correlations for multi-label classification exploited graphical models such as Conditional Random Fields (CRFs)~\cite{ghamrawi2005collective} or Dependency Networks~\cite{guo2011multi}. More recently, the idea to exploit RNN models to capture label correlations was proposed in~\cite{jin2016annotation} and~\cite{wang2016cnn}, where the low dimensional internal state of the network was used to model label dependencies. Wang et al.~\cite{wang2016cnn} combined CNN and RNN architectures, and learned a joint image-label embedding space to learn label dependencies. However, since LSTMs produce sequential outputs, a fixed order was imposed to the labels, and the model learned to predict the output in the same order. This led to problems such as skipping predictions for classes that appear earlier in the sequence but less relevant in the image. Yazici et al.~\cite{yazici2020orderless}, proposed a CNN-RNN model which they trained with an orderless loss function to avoid the drawbacks of imposing a fixed label order in RNN. Finally, ML-GCN~\cite{chen2019multi}, exploited graph convolutional networks to capture label dependencies.

\minisection{Mixup.}
Zhang et al.~\cite{zhang2017mixup} proposed to blend images and their associated labels randomly to improve generalization of the models. It was shown that the mixup was beneficial to avoid overconfident predictions in several tasks such as image classification~\cite{zhang2017mixup, thulasidasan2019mixup}, object detection~\cite{zhang2019bag}, text classification~\cite{thulasidasan2019mixup} and semantic segmentation~\cite{islam2020feature}. Verma et al.~\cite{verma2019manifold} proposed to combine hidden states of paired samples in addition to their images and labels. Islam et al.~\cite{islam2020feature} conditioned the mixup on the labels of paired samples. If the mixup is done without any constraints, then the majority of the pairs will mostly include the most frequent classes. To avoid the model to have a strong bias for frequent classes, they combined images based on a uniform distribution across categories. Wang et al.~\cite{wang2019baseline} is the first work that exploited the mixup technique for multi-label image classification. Although the authors did not report any improvement over the baselines in case of single models, they noted that an ensemble of models trained with mixup achieved better results.

\section{Method}
Our method for the multi-label classification is based on a transformer architecture. We try different strategies to assign labels to the decoder output. In this section, we briefly introduce some transformer concepts, explain our proposed architecture with different losses and, finally, present our different adaptations of the mixup technique.

\subsection{Transformers}
The main component of transformers is a self-attention module~\cite{vaswani2017attention} which uses a set of weights $W$ to compute the \emph{query} ($Q$), \emph{key} ($K$) and \emph{value} ($V$) vectors with the input vector:
\begin{equation}
\begin{aligned}
Q = W_{Q}X, \quad K = W_{K}X, \quad V = W_{V}X
\end{aligned}
\end{equation}
Then, these three vectors are combined to compute the output of the self-attention layer:
\begin{equation}
\begin{aligned}
\textrm{A} = \textrm{softmax}(\frac{QK^{T}}{\sqrt{D}}) V
\end{aligned} \label{eq:attention}
\end{equation}
The result of the dot product between the \emph{query} and the \emph{key} is divided by the square root of the dimensionality $D$ to have smoother softmax values.
Many encoder layers, consisting of a self-attention and a feed-forward neural network are stacked in the encoder part of the network, and the output of the final encoder layer ($A$) is passed to the \emph{cross attention module} in every decoder layer. The cross attention module has the same structure as the self attention module, with the only difference that the output of the final encoder layer is used as the key and value vectors for every decoder layer during the forward pass, while the query is obtained from the input of the decoder.

The input of the transformer decoder stack depends on the task and design of the architecture. In case of an NLP task, it might be class embeddings~\cite{zhu2018captioning} or masked output embeddings~\cite{vaswani2017attention}. In case of a vision-based task, it might be a set of object queries~\cite{carion2020end}. In~\cite{carion2020end}, the object queries are added to the query input of each decoder layer (the query input of the first decoder layer consists of zeros as can be seen in Figure~\ref{fig:our_detr_model}). The fact that object queries are provided to all decoder layers complicates their training: we conjecture that the requirement to be useful at multiple hierarchical levels (and therefore at different semantic levels) is hard to fulfill. In order to verify this, we conduct several experiments where we compare the approach in~\cite{carion2020end} (denominated as DETR) with inputting unique sets of object queries to each decoder layer (denominated as DETR*). The first two rows of Table~\ref{tab:ours_detr}, show the performance of the two models with different number of decoder layers. DETR* obtains significantly better results than DETR which confirms our assumption that inputting the same set of object queries to different decoder layers is not optimal. However, although learning a separate set of object queries for each decoder layer improves the results, it might be computationally redundant and not feasible when the number of decoder layers increases.

\begin{table}[!tb]
\centering
\caption{\small Comparison of the performances of DETR, DETR* and T-POQ models with different number of decoder layers on MS-COCO.}
\scalebox{0.85}{\begin{tabular}{c|ccccccccc}
      & \multicolumn{9}{c}{\# of decoder layers}                                                                                                                                                                                            \\
      & \multicolumn{3}{c}{2}                                                                    & \multicolumn{3}{c}{3}                                                                    & \multicolumn{3}{c}{4}                         \\ \hline
      & C-P                      & C-R                      & \multicolumn{1}{c|}{C-F1}          & C-P                      & C-R                      & \multicolumn{1}{c|}{C-F1}          & C-P           & C-R           & C-F1          \\ \hline
DETR  & 75.9                     & 64.7                     & \multicolumn{1}{c|}{69.9}          & 74.9                     & 64.6                     & \multicolumn{1}{c|}{69.4}          & \textbf{76.8} & 64.2          & 69.9          \\
DETR* & \multicolumn{1}{r}{75.9} & \multicolumn{1}{r}{65.9} & \multicolumn{1}{r|}{70.5}          & \multicolumn{1}{r}{75.9} & \multicolumn{1}{r}{66.1} & \multicolumn{1}{r|}{70.7}          & 76.5          & 65.4          & \textbf{70.5} \\
T-POQ  & \textbf{76.6}            & \textbf{66.0}            & \multicolumn{1}{c|}{\textbf{70.9}} & \textbf{76.5}            & \textbf{66.1}            & \multicolumn{1}{c|}{\textbf{70.9}} & 73.8          & \textbf{67.1} & 70.3          
\end{tabular}}
\label{tab:ours_detr}
\end{table}

\subsection{Overview of architecture}
\begin{figure}[!tb]
\centering
  \includegraphics[width=0.8\linewidth]{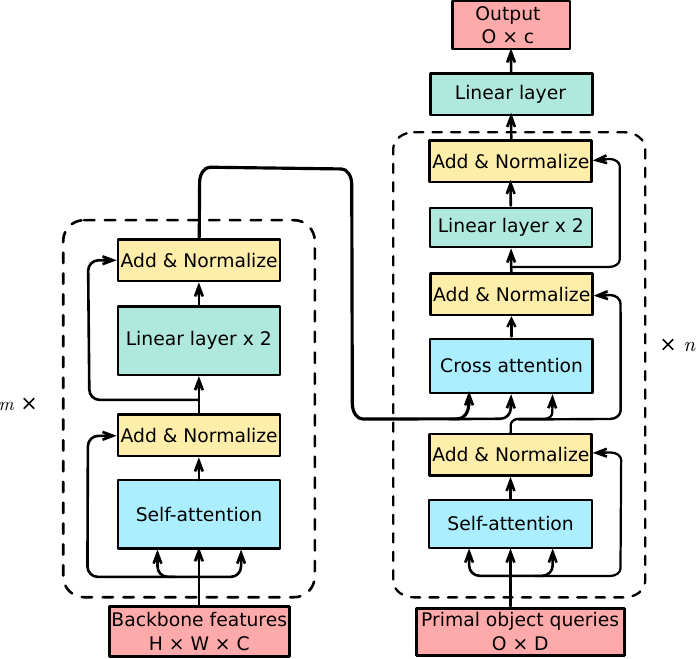}
  \caption{\small The overall transformer architecture used in this paper.}
  \label{fig:architecture}
\end{figure}

Our architecture consists of two parts: a backbone that processes the input image, and a transformer, which can be further divided into encoder and decoder (see Figure~\ref{fig:architecture}).

\paragraph{Backbone} We use a convolutional neural network to obtain a feature representation of the input image, but since the transformer requires sequential data, we linearize the feature map of the last convolutional layer of the backbone network along the spatial dimensions, and use them as the input sequence for the encoder. More precisely, let the last convolutional layer of the CNN output a feature map of size $H \times W \times C$, where $H$ and $W$ are the spatial dimensions, and $C$ the number of channels, we reshape it to $ HW \times C$ to obtain a sequence of feature vectors with dimension $C$.

\paragraph{Transformer encoder and decoder} In the transformer encoder, the linearized feature map is passed through the self-attention and linear layers, and then the result is passed to the cross attention module of every decoder layer. The initial linear layer reduces the dimension to $D$. We do not add learnable or fixed positional encodings to encoder features since it does not give any improvements. We attribute this to using semantically strong features that do not require additional spatial information for the task of multi-label classification. We quantitatively verified that adding positional encodings is only beneficial when features from earlier layers are used, however, that also leads to lower results.

The input of the first decoder layer are \emph{primal object queries} which differ from the object queries proposed in~\cite{carion2020end} in the way that they are given to transformer decoder stack. In our model, primal object queries are given directly to the first decoder layer instead of being added as positional encodings in every decoder layer. In this way, we avoid the drawbacks of the standard object queries which were mentioned in the previous section. In Table~\ref{tab:ours_detr}, we also added the results of our method (T-POQ) and it can be seen that without the overhead of learning separate object queries for each decoder layer (as does DETR*) our method obtains equal or better results. Interestingly, in the experimental section we will also show that T-POQ obtains a significant speed-up in training when compared to the DETR model. Finally, the number of primal object queries is $O$ while the dimension of the queries is $D$. Then, a linear layer outputs a tensor whose shape is $O \times c$ where $c$ is the number of classes.

\paragraph{Training losses}
Using object queries for multi-label classification is a new concept. Previously, e.g. in CNN-RNN models, the labels are ordered either dynamically~\cite{yazici2020orderless} or an imposed fixed-order is applied~\cite{wang2016cnn}. The forward pass is done in a recursive way, therefore the length of the output tensor of the RNN is bounded by the number of labels of an image that has the most labels. However, in case of transformers, the forward pass is not done recursively which gives the flexibility of determining the size of the output tensor. We will consider two ways to train transformers for multi-label classification. 

Firstly, we consider the case where we align each object query with a specific class. In this case $O=c$ and each object query specializes in detecting a single object class. In case of non-existent labels, \emph{empty} tokens are assigned to the object queries that are in charge of these labels. We call this the \emph{exhaustive model}. Here is the loss equation for the exhaustive model: 
\begin{equation}
\begin{aligned}
& S_{ji} = \frac{e^{x_{ji}}}{\sum_{i=1}^c e^{x_{ji}}}, \quad \mathcal{L}_{exh} = -\frac{1}{O}\sum_{j=1}^{O}\sum_{i=1}^{c} y_{ji}\log S_{ji} \\[-2pt]
& \text{subject to}
\quad y_{ji} \in \{0, 1\},\\[-2pt]
& \quad \sum\nolimits_j {y_{ji}  = 1}\;\;\forall\;i\in L \quad\text{and} \quad \sum\nolimits_j {y_{ji}  = 0}\;\;\forall\;i\notin L
\end{aligned}
\end{equation}
where $y_{ji} \in \mathbb{N}^{O\times c}$, $x_{ji}$ and $L$  are the class labels, output tensor and set of class labels, respectively. A drawback of the exhaustive approach is that it scales linearly with the number of classes, and might be infeasible for datasets with many labels. However, reducing the number of object queries requires one object query to be in charge  of  multiple  labels,  which results in an assignment problem. Therefore, we employ an orderless loss inspired by the one proposed in~\cite{yazici2020orderless}. We call this model the \emph{aligned model} and it no longer requires to scale linearly with the number of classes in the dataset:
\begin{equation}
\begin{aligned}
 & \mathcal{L}_{align} = \mathop{\min}\limits_y \sum_{i=1}^{c} y_{ji}\log S_{ji} \\[-2pt]
 & \text{subject to} \quad y_{ji} \in \{ 0,1\}, y_{ji} = 1\;\text{if} \;\; \hat{l}_j \in L \;\text{and} \; i=\hat{l}_j , \\[-2pt]
 & \quad \sum\nolimits_j {y_{ji}  \geq 1}\;\;\forall\;i\in L \quad \sum\nolimits_j {y_{ji}  = 0}\;\;\forall\;i\notin L
 \end{aligned}\label{eq:PLA}
\end{equation}
where $\hat{l}_j$ is the class predicted by the model at object query $j$. The order of the labels in $y$ is chosen in such a way that it gives the minimum cross entropy loss. This label assignment problem can be solved with the Hungarian algorithm. In addition, we impose the constraint, which was proposed in~\cite{yazici2020orderless}, that assigns the class $\hat{l}_j$ to object query $j$ if $\hat{l}_j$ belongs to $L$. Therefore, the same class may be assigned to several object queries. 

\subsection{Mixup}
Mixup was proposed in~\cite{zhang2017mixup} and was found to significantly improve results for image classification, object detection and NLP tasks~\cite{zhang2017mixup, thulasidasan2019mixup, zhang2019bag}. We consider three ways of adapting the mixup technique to the training: soft, hard and restricted hard.

To train the model, a dataset with pairs of images and sets of labels is used. Let $\left(I, T\right)$ be the pairs in a batch containing $N$ images $I = \{i_1, i_2, ..., i_{N}\}$ and labels $T=\{t_1,t_2,...,t_{N}\}, \; t_i \in \{0,1\}^c$ where $c$ is the number of classes in the dataset. For the \emph{soft mixup}, which is the original mixup as proposed in~\cite{zhang2017mixup}, we sample random weights from a beta distribution $\lambda \sim Beta(\alpha, \alpha), \alpha \in (0, \infty)$ to use them to mixup images and their associated labels: 
\begin{equation}
\begin{aligned}
i_m = \lambda i_i + (1 - \lambda) i_j \\
t_m = \lambda t_i + (1 - \lambda) t_j \\
\end{aligned}
\end{equation}
where $i_i$ and $i_j$ are randomly selected images, and $i_m$ is the mixed up version. 

The soft mixup makes sense for single class image classification where the last layer is typically a softmax. However, for multi-label image classification multiple labels can be present in the image. Therefore, we use the mixup proposed in~\cite{wang2019baseline} and denominate it as \emph{hard mixup} where the union of labels is taken instead of the average. As proposed by the author, we alternatively enable and disable the application of mixup in every epoch and we use the ratio of $0.5 : 0.5$ for images.
In order to apply the mixup in all epochs and have both mixed and non-mixed images in a batch we consider the \emph{restricted hard mixup} as a final setup, where we apply the mixup in every epoch and restrict it by applying it only to half of the images in the batch. We use the last half of the batch to mix with the first half.
For instance, if we set the batch size to 4, after the mixup the images and labels become $I = \{(i_1+i_3) / 2, (i_2+i_4) / 2, i_3, i_4\}$ and $T=\{t_1 \cup t_3, t_2 \cup t_4, t_3, t_4\}$ respectively.

\section{Experiments}
\minisection{Datasets and setting.}
We evaluate our models on MS-COCO \cite{lin2014microsoft} and NUS-WIDE\cite{chua2009nus} datasets. \textbf{MS-COCO} consists of 82,081 training and 40,137 test images for 80 object categories. \textbf{NUS-WIDE} consists of 269,648 images with a total number of 5,018 unique labels. However, annotations for 81 labels are more trustworthy and used for evaluation. After removing images that do not belong to the 81 labels, 209,347 images remain.

\minisection{Evaluation metrics.}
We use \textit{per-class} and \textit{overall} precision, recall and F1 scores. For the MS-COCO, we also report the mean average precision (mAP) score.

\minisection{Network training.}
The number of object queries is set to 25 for all datasets. The internal dimension of the transformer is 512. The backbone network, which is pre-trained on ImageNet~\cite{russakovsky2015imagenet}, uses an SGD optimizer with learning rate $0.001$ and momentum $0.9$. The rest of the model is trained with the ADAM optimizer with a learning rate $0.0001$ for 40 epochs. The batch size is 32, and random affine transformations and contrast changes are applied as data augmentation. The batch norm layers in the backbone are frozen during the training\footnote{ \url{https://github.com/voyazici/visual-transformers-classification}}.

\subsection{Ablation}
All the ablation studies are done on MS-COCO dataset. and are repeated three times. The final result is the average of the three experiments. For validation we use 25\% of training data and the rest as train set. The snapshot that gives the highest score on the validation set is evaluated on the test set. The input image size is 288. Unless stated otherwise, the used transformer model is the aligned model.

In Table~\ref{tab:ablation1}, LSTM and transformer models with different encoder and decoder layers are compared. The LSTM model is trained with an orderless loss~\cite{yazici2020orderless} and has the same backbone and hidden size as the transformer model.  From the results it can be seen that adding encoder layers does not yield any significant improvement. When we evaluate the attention maps generated by the self-attention module in the encoder, unlike in~\cite{carion2020end}, the  encoder does not separate instances which is not crucial for multi-label classification unlike for object-detection.  On the other hand, more decoder layers do lead to improvement in the recall metric.
Moreover, our transformer model slightly increases the total number of parameters and computational cost ($55.2$M and $27.1$ GFLOPS) compared to the backbone model ($44.6$M and $25.9$ GFLOPS).

\begin{table}[htbp]
\centering
\caption{\small Comparison of LSTM and different transformer models.}
\scalebox{0.8}{\begin{tabular}{c|c|c|r|rrr|rrr}
Model  & \# of enc. & \# of dec. & \multicolumn{1}{c|}{mAP}  & \multicolumn{1}{c}{C-P}  & \multicolumn{1}{c}{C-R}  & \multicolumn{1}{c|}{C-F1} & \multicolumn{1}{c}{O-P}  & \multicolumn{1}{c}{O-R}  & \multicolumn{1}{c}{O-F1} \\ \hline
LSTM   & -          & -          & \multicolumn{1}{c|}{75.3} & \multicolumn{1}{c}{76.2} & \multicolumn{1}{c}{66.7} & \multicolumn{1}{c|}{71.1} & \multicolumn{1}{c}{78.8} & \multicolumn{1}{c}{71.3} & \multicolumn{1}{c}{74.9} \\
Trans. & 0          & 1          & 78.0                      & 79.6                     & 67.9                     & 73.3                      & \textbf{81.8}            & 71.9                     & 76.5                     \\
Trans. & 1          & 1          & 77.6                      & 77.9                     & 68.8                     & 73.1                      & 80.2                     & 72.6                     & 76.2                     \\
Trans. & 1          & 2          & \textbf{78.3}             & \textbf{79.6}            & 68.2                     & \textbf{73.5}             & 80.8                     & 72.4                     & 76.3                     \\
Trans. & 2          & 2          & 78.2                      & 78.2                     & 69.3                     & 73.5                      & 80.2                     & 73.1                     & \textbf{76.5}            \\
Trans. & 2          & 3          & 77.9                      & 77.4                     & \textbf{69.6}            & 73.3                      & 79.3                     & \textbf{73.6}            & 76.4                    
\end{tabular}}
\label{tab:ablation1}
\end{table}

\begin{table}[htbp]
\centering
\caption{\small Subtraction of LSTM scores from transformer scores on different number of labels per image.} 
\scalebox{0.8}{
\begin{tabular}{c|cccccccccc|c}
          & \multicolumn{10}{c|}{Number of labels per image}           &     \\ \hline
          & 1   & 2   & 3   & 4   & 5   & 6   & 7   & 8    & 9   & +10 & Avg \\ \cline{2-12} 
precision & 2.1 & 1.5 & 1.5 & 2.4 & 1.3 & 1.8 & 1   & -0.5 & 0.7 & 0.0 & 0.8 \\
recall    & 0.5 & 1   & 1.1 & 1.7 & 1.2 & 1.5 & 1.2 & 2    & 2.5 & 2.1 & 1.7 \\
F1        & 1.5 & 1.3 & 1.3 & 2   & 1.3 & 1.6 & 1.1 & 1.1  & 1.9 & 1.5 & 1.5
\end{tabular}}
\label{tab:trans_lstm}
\end{table}

In Table~\ref{tab:trans_lstm}, we compare the performance of the LSTM and transformer model (the one with one encoder and two decoder layers) on images that have different number of labels. The values are the subtraction of average LSTM scores from average transformer scores. When the number of labels increases, the transformer model misses fewer classes that leads to fewer false negatives and higher recall.

\begin{table}[b]
\centering
\caption{\small Comparison of exhaustive and aligned models and mixup methods.}
\scalebox{0.8}{
\begin{tabular}{c|c|ccc|ccc}
                     & mAP                                & C-P                               & C-R                               & C-F1                               & O-P                               & O-R                                  & O-F1                              \\ \hline
Exhaustive           & 78.3                               & 77.5                              & 69.5                              & 73.2                               & 79.8                              & 73.3                              & 76.4                              \\
Aligned              & 78.1                               & 77.6                              & 69.4                              & 73.3                               & 79.9                              & 73.3                              & 76.4                              \\ \hline
Aligned + soft mixup & \multicolumn{1}{l|}{78.6}          & \multicolumn{1}{l}{79.1}          & \multicolumn{1}{l}{\textbf{69.8}} & \multicolumn{1}{l|}{74.2}          & \multicolumn{1}{l}{80.9}          & \multicolumn{1}{l}{\textbf{73.7}} & \multicolumn{1}{l}{77.1}          \\
Aligned + hard mixup & \multicolumn{1}{l|}{79.0} & \multicolumn{1}{l}{79.7} & \multicolumn{1}{l}{69.4}          & \multicolumn{1}{l|}{74.2} & \multicolumn{1}{l}{81.4} & \multicolumn{1}{l}{73.3}          & \multicolumn{1}{l}{77.1} \\
Aligned + restr. hard mixup & \multicolumn{1}{l|}{\textbf{79.6}} & \multicolumn{1}{l}{\textbf{80.2}} & \multicolumn{1}{l}{69.7}          & \multicolumn{1}{l|}{\textbf{74.6}} & \multicolumn{1}{l}{\textbf{82.1}} & \multicolumn{1}{l}{73.5}          & \multicolumn{1}{l}{\textbf{77.5}} \\

\end{tabular}}
\label{tab:ablation2}
\end{table}

In Table~\ref{tab:ablation2}, aligned and exhaustive models (one encoder and two decoder layers) are compared. The results of the models are comparable. However, for the comparison with the state-of-the-art models, we will use the aligned model. It is a more compact model and the overhead cost of the alignment step (0.9 ms per image) is negligible.
Also in Table~\ref{tab:ablation2}, different mixup setups are compared. The $\alpha$ value for the soft mixup is $0.4$. 
Mixup improves the results considerably. The best result is obtained with the restricted hard mixup. Higher precision and mAP scores show that the restricted hard mixup model has more confident predictions. We attribute the superiority of the restricted hard mixup over the soft mixup to soft labels being detrimental for modelling label correlations; and the superiority over the hard mixup to lower variance during the gradient update due to having mixed and non-mixed samples in the same batch.

\subsection{Object Queries}

\begin{figure}[!tb]
\centering
  \includegraphics[width=0.8\linewidth]{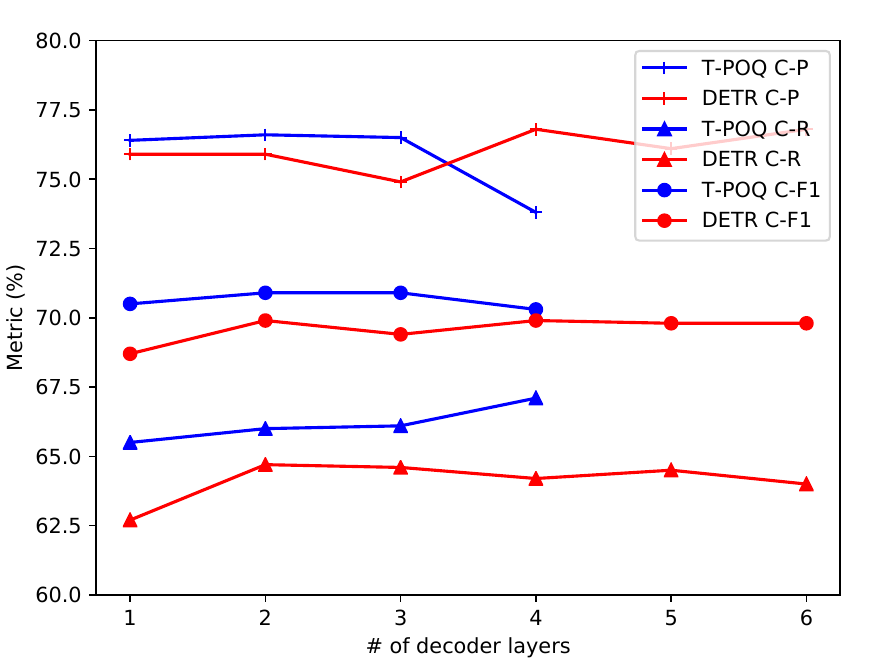}
  \caption{\small Comparison of the proposed and DETR approaches with different number of decoder layers on MS-COCO.}
  \label{fig:compare_query}
\end{figure}

\begin{figure}[!tb]
\centering
  \includegraphics[width=0.8\linewidth]{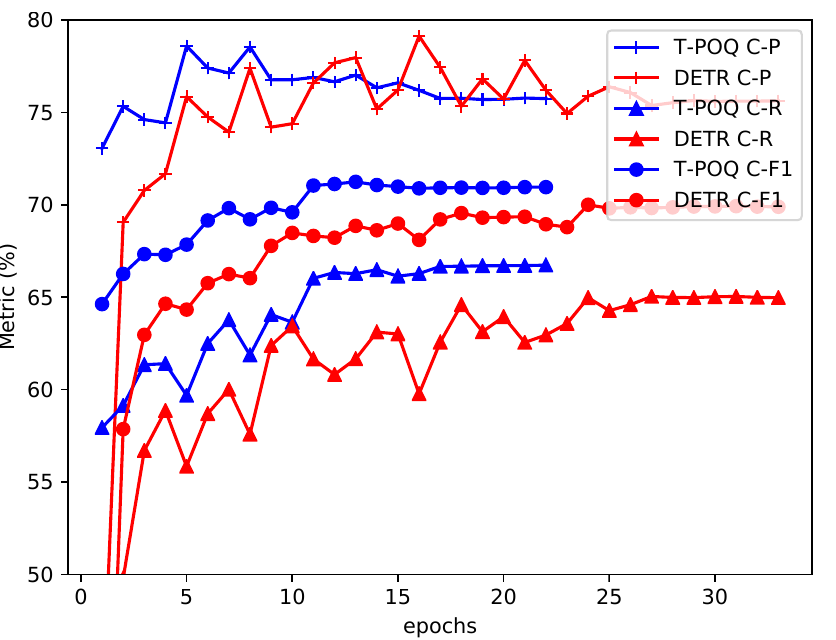}
  \caption{\small Convergence of the proposed and DETR models on MS-COCO.}
  \label{fig:compare_query_2}
\end{figure}

In Figure~\ref{fig:compare_query}, we compare our primal object queries with the object queries in DETR~\cite{carion2020end}. We show the change of C-P, C-R and C-F1 metrics with different number of decoder layers. For simplicity, we do not employ the mixup and the image size is $224 \times 224$. We set the learning rate of the backbone to 0.0001 for the DETR model to make the model converge during training. It can be seen that our approach  yields significantly higher C-F1 scores in every setup. Also, with more decoder layers it achieves higher recall. In addition, the model with one decoder layer already performs comparable with the models that have more decoder layers. On the other hand, the model with the DETR approach requires at least two decoder layers to achieve comparable performance. We attribute this superiority to the residual connection after the cross attention module which enables the propagation of our primal object queries to the next layer. We empirically confirm this by disabling the residual connection and obtaining the same results as the DETR model in the case that the number of decoder layers is one. The same fact causes 79.0\% and 38.6\% faster convergence on MS-COCO and NUS-WIDE datasets respectively. In order to compare the convergence between the two models, we determine the epoch number at which C-F1 stops improving for both models. Then, we calculate the percentage change of the epoch number for setups that have up to three decoder layers. Finally, we average the percentages from different setups to get the relative change in the convergence speed. In Figure~\ref{fig:compare_query_2}, the DETR model starts to converge much later than our model (the setup with three decoder layers). Moreover, due to the lack of the propagation of the object queries, the DETR model starts from a significantly lower point compared to our model. Consequently, when we compare the best models, we achieve improvements over the DETR approach by 1.2\%-1.5\% in all metrics.

\begin{figure}[!htbp]
  \centering
  \includegraphics[width=0.9\linewidth]{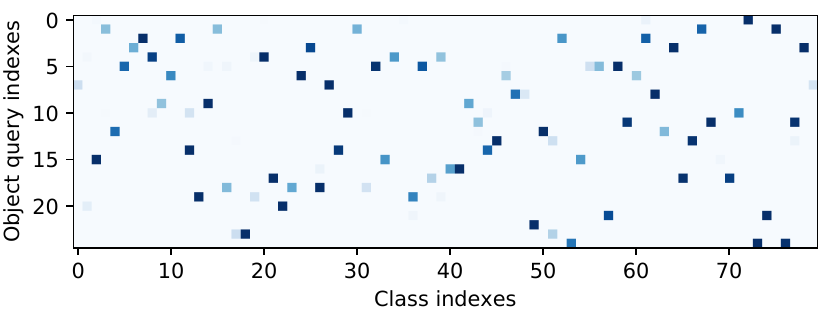}
  \caption{\small Normalized counts of predicted classes. Each object query recognizes a subset of classes.} 
  \label{fig:object_query}
\end{figure}

Next, we analyze what the primal object queries of the aligned model learn when they must account for multiple classes with the orderless loss. In Figure~\ref{fig:object_query} (darker shades indicate higher values), we display the normalized counts of the predicted classes for each object query. Each object query learns to recognize a subset of classes. Since each of them can only make one prediction during one forward pass, the subsets consist of classes that are not likely to exist together. For example, the fourth object query learns to recognize \textit{bear}, \textit{cow}, \textit{suitcase}, and \textit{wine glass}.

\subsection{Comparison with the SOTA}
We compare our results with several recent models: CNN-RNN~\cite{wang2016cnn}, SR CNN-RNN~\cite{liu2017semantic}, Chen et al.~\cite{chen2018order}, Li et al.~\cite{li2018attentive} and PLA~\cite{yazici2020orderless} are models that include RNNs either to model label relations or recursively generate attention maps. SRN~\cite{zhu2017learning} proposes a Spatial Regularization Network that generates attention maps for all labels. ACfs~\cite{guo2019visual}, proposes a two-branch network with an original image and its transformed image as inputs and imposes an additional loss to ensure the consistency between attention maps of both versions. DER~\cite{chen2020disentangling} proposes to train a detection model in three steps in which it learns class-aware attention maps, models label correlations explicitly and measures similarity between label embeddings. ML-GCN~\cite{chen2019multi}, exploits graph convolutional networks to capture label dependencies. C-Tran~\cite{lanchantin2021general} is the only transformer model that we compare with. It exploits self-attention layers in a transformer encoder to learn label correlations given image features and a set of masked label embeddings and does not include any transformer decoder layer unlike our model.
For MS-COCO experiments ResNet-101 architecture is used for the backbone, and for NUS-WIDE we run experiment with both ResNet101 and VGG16. For the comparison with SOTA models, we use the aligned transformer model with one encoder and two decoder layers. We also employ the restricted hard mixup.

We outperform all the state-of-art on MS-COCO (see Table~\ref{tab:sota_coco}).
The performance superiority is more apparent in the recall metrics, since the transformer model is less likely to miss classes. The results on the NUS-WIDE dataset can be seen in Table~\ref{tab:sota_nus}. The results on the top part of the table use the split proposed by~\cite{johnson2015love} with a VGG16 backbone, while the ones on the lower part use the original split and a ResNet-101 backbone. For the T-POQ* model, we resize the input image to $448 \times 448$ to make the comparison fair with the DER~\cite{chen2020disentangling}. For the T-POQ model, we resize it to $224 \times 224$ to make the comparison fair with SRN~\cite{zhu2017learning}. We surpass all other models, especially in the class-wise metrics which are more relevant, since NUS-WIDE is an unbalanced dataset.

\begin{table}[tbp]
\centering
\caption{\small Comparison with state-of-the-art on MS-COCO.}
\scalebox{0.85}{
\begin{tabular}{c|c|c|ccc|ccc}
\hline
Methods & \begin{tabular}[c]{@{}c@{}}Image\\ size\end{tabular} & mAP  & C-P  & C-R  & C-F1 & O-P  & O-R  & O-F1 \\ \hline
SRN~\cite{zhu2017learning}     & 224                                                        & 77.1 & 81.6 & 65.4 & 71.2 & 82.7 & 69.9 & 75.8 \\
ACFs~\cite{guo2019visual}    & 288                                                        &   77.5   & 77.4 & 68.3 & 72.2 & 79.8 & 73.1 & 76.3 \\
PLA~\cite{yazici2020orderless}     & 288                                                        & -    & 80.4 & 68.9 & 74.2 & 81.5 & 73.3 & 77.1 \\
ML-GCN~\cite{chen2019multi}  & 448                                                        & 83.0 & 85.1 & 72.0 & 78.0 & 85.8 & 75.4 & 80.3 \\
DER~\cite{chen2020disentangling}    & 448                                                        & 82.9 & 84.7 & 71.6 & 77.6 & 86.0 & 74.9 & 80.0 \\ 
C-Tran~\cite{lanchantin2021general} & 576 & 85.1 & \textbf{86.3} & 74.3 & 79.9 & \textbf{87.7} & 76.5 & 81.7 \\ \hline 
T-POQ  & 224                                                        &  77.9    &  79.5    &    67.4  & 73.0     &  81.5    &   71.3   &    76.1  \\
T-POQ   & 288                                                        &   80.6   & 80.9 & 70.9 & 75.6 & 82.5 & 74.4 & 78.2 \\
T-POQ   & 448                                                        &  84.5    &  82.9    & 75.8   & 79.2     &    84.4 &  78.4    &    81.3 \\
T-POQ   & 576                                                        &  \textbf{86.2}    &  84.1    & \textbf{77.9}   & \textbf{80.9}     &    85.0 &  \textbf{80.6}    &    \textbf{82.8} \\ \hline

\end{tabular}}
\label{tab:sota_coco}
\end{table}

\begin{table}[tbp]
\centering
\caption{\small Comparison with state-of-the-art on NUS-WIDE.}
\scalebox{0.85}{
\begin{tabular}{c|ccc|ccc}
\hline
Methods             & C-P & C-R & C-F1  & O-P & O-R & O-F1  \\ \hline
CNN-RNN \cite{wang2016cnn} & 40.5 & 30.4 & 34.7 & 49.9 & 61.7 & 55.2\\ 
Chen et al.~\cite{chen2018order}   & 59.4 & 50.7 & 54.7 & 69.0 & 71.4 & 70.2 \\
SR CNN-RNN \cite{liu2017semantic}  & 55.7 & 50.2 & 52.8 & 70.6 & 71.4 & 71.0 \\
Li et al.~\cite{li2018attentive}  & 44.2 & 49.3 & 46.6 & 53.9 & 68.7 & 60.4  \\
LSEP \cite{li2017improving} & \textbf{66.7} & 45.9 & 54.4 & \textbf{76.8} & 65.7 & 70.8 \\
PLA~\cite{yazici2020orderless} & 60.7 & 52.4 & 56.2 & 72.0 & \textbf{72.8} & 72.4 \\\hline
T-POQ & 66.0 & \textbf{52.7} & \textbf{58.6} & 74.7 & 71.8 & \textbf{73.2} \\ \hline \hline
SRN~\cite{zhu2017learning} & 65.2 & 55.8 & 58.5 & 75.5 & 71.5 & 73.4 \\ 
DER~\cite{chen2020disentangling} & 64.2 & \textbf{57.9} & 60.9 & 75.5 & 73.0 & 74.2 \\ \hline
T-POQ & 66.5 & 56.0 & 60.8 & 75.1 & 73.2 & 74.1 \\
T-POQ* & \textbf{66.8} & 57.8 & \textbf{62.0} & \textbf{75.6} & \textbf{74.3} & \textbf{74.9} \\ \hline

\end{tabular}}
\label{tab:sota_nus}
\end{table}

\section{Conclusions}
We introduced the \emph{primal object queries} that achieved significantly better results and a large speed-up of training convergence for both MS-COCO and NUS-WIDE datasets. Our model is unique in that it achieves to learn long-term dependencies, adapts and integrates the mixup technique for multi-label classification successfully, and obtains state-of-the-art results for multi-label classification on the MS-COCO and NUS-WIDE datasets by a large margin.

\minisection{Acknowledgements:} We acknowledge the Spanish projects PID2019-104174GB-I00 and the Industrial Doctorate Grant 2016 DI 039 of the Ministry of Economy and Knowledge of the Generalitat de Catalunya, and its CERCA Program.

\bibliographystyle{IEEEtran}
\bibliography{IEEEabrv,egbib}

\begin{thebibliography}{10}
\providecommand{\url}[1]{#1}
\csname url@samestyle\endcsname
\providecommand{\newblock}{\relax}
\providecommand{\bibinfo}[2]{#2}
\providecommand{\BIBentrySTDinterwordspacing}{\spaceskip=0pt\relax}
\providecommand{\BIBentryALTinterwordstretchfactor}{4}
\providecommand{\BIBentryALTinterwordspacing}{\spaceskip=\fontdimen2\font plus
\BIBentryALTinterwordstretchfactor\fontdimen3\font minus
  \fontdimen4\font\relax}
\providecommand{\BIBforeignlanguage}[2]{{%
\expandafter\ifx\csname l@#1\endcsname\relax
\typeout{** WARNING: IEEEtran.bst: No hyphenation pattern has been}%
\typeout{** loaded for the language `#1'. Using the pattern for}%
\typeout{** the default language instead.}%
\else
\language=\csname l@#1\endcsname
\fi
#2}}
\providecommand{\BIBdecl}{\relax}
\BIBdecl

\bibitem{bi2013efficient}
W.~Bi and J.~Kwok, ``Efficient multi-label classification with many labels,''
  in \emph{International Conference on Machine Learning}.\hskip 1em plus 0.5em
  minus 0.4em\relax PMLR, 2013, pp. 405--413.

\bibitem{wehrmann2018hierarchical}
J.~Wehrmann, R.~Cerri, and R.~Barros, ``Hierarchical multi-label classification
  networks,'' in \emph{International Conference on Machine Learning}.\hskip 1em
  plus 0.5em minus 0.4em\relax PMLR, 2018, pp. 5075--5084.

\bibitem{ghamrawi2005collective}
N.~Ghamrawi and A.~McCallum, ``Collective multi-label classification,'' in
  \emph{Proceedings of the 14th ACM International Conference on Information and
  Knowledge Management}.\hskip 1em plus 0.5em minus 0.4em\relax ACM, 2005, pp.
  195--200.

\bibitem{guo2011multi}
Y.~Guo and S.~Gu, ``Multi-label classification using conditional dependency
  networks,'' in \emph{Twenty-Second International Joint Conference on
  Artificial Intelligence}, 2011.

\bibitem{wang2016cnn}
J.~Wang, Y.~Yang, J.~Mao, Z.~Huang, C.~Huang, and W.~Xu, ``Cnn-rnn: A unified
  framework for multi-label image classification,'' in \emph{Proceedings of the
  IEEE/CVF Conference on Computer Vision and Pattern Recognition}, 2016, pp.
  2285--2294.

\bibitem{liu2017semantic}
F.~Liu, T.~Xiang, T.~M. Hospedales, W.~Yang, and C.~Sun, ``Semantic
  regularisation for recurrent image annotation,'' in \emph{Proceedings of the
  IEEE/CVF Conference on Computer Vision and Pattern Recognition}, 2017, pp.
  2872--2880.

\bibitem{yazici2020orderless}
V.~O. Yazici, A.~Gonzalez-Garcia, A.~Ramisa, B.~Twardowski, and J.~v.~d.
  Weijer, ``Orderless recurrent models for multi-label classification,'' in
  \emph{Proceedings of the IEEE/CVF Conference on Computer Vision and Pattern
  Recognition}, 2020, pp. 13\,440--13\,449.

\bibitem{guo2019visual}
H.~Guo, K.~Zheng, X.~Fan, H.~Yu, and S.~Wang, ``Visual attention consistency
  under image transforms for multi-label image classification,'' in
  \emph{Proceedings of the IEEE/CVF Conference on Computer Vision and Pattern
  Recognition}, 2019, pp. 729--739.

\bibitem{vaswani2017attention}
A.~Vaswani, N.~Shazeer, N.~Parmar, J.~Uszkoreit, L.~Jones, A.~N. Gomez,
  {\L}.~Kaiser, and I.~Polosukhin, ``Attention is all you need,'' in
  \emph{Advances in Neural Information Processing Systems}, 2017, pp.
  5998--6008.

\bibitem{devlin-etal-2019-bert}
J.~Devlin, M.-W. Chang, K.~Lee, and K.~Toutanova, ``{BERT}: Pre-training of
  deep bidirectional transformers for language understanding,'' in
  \emph{Proceedings of the 2019 Conference of the North {A}merican Chapter of
  the Association for Computational Linguistics: Human Language Technologies,
  Volume 1 (Long and Short Papers)}.\hskip 1em plus 0.5em minus 0.4em\relax
  Minneapolis, Minnesota: Association for Computational Linguistics, Jun. 2019,
  pp. 4171--4186.

\bibitem{cornia2020meshed}
M.~Cornia, M.~Stefanini, L.~Baraldi, and R.~Cucchiara, ``Meshed-memory
  transformer for image captioning,'' in \emph{Proceedings of the IEEE/CVF
  Conference on Computer Vision and Pattern Recognition}, 2020, pp.
  10\,578--10\,587.

\bibitem{zhu2018captioning}
X.~Zhu, L.~Li, J.~Liu, H.~Peng, and X.~Niu, ``Captioning transformer with
  stacked attention modules,'' \emph{Applied Sciences}, vol.~8, no.~5, p. 739,
  2018.

\bibitem{carion2020end}
N.~Carion, F.~Massa, G.~Synnaeve, N.~Usunier, A.~Kirillov, and S.~Zagoruyko,
  ``End-to-end object detection with transformers,'' in \emph{European
  Conference on Computer Vision}.\hskip 1em plus 0.5em minus 0.4em\relax
  Springer, 2020, pp. 213--229.

\bibitem{dosovitskiy2021an}
A.~Dosovitskiy, L.~Beyer, A.~Kolesnikov, D.~Weissenborn, X.~Zhai,
  T.~Unterthiner, M.~Dehghani, M.~Minderer, G.~Heigold, S.~Gelly, J.~Uszkoreit,
  and N.~Houlsby, ``An image is worth 16x16 words: Transformers for image
  recognition at scale,'' in \emph{International Conference on Learning
  Representations}, 2021.

\bibitem{lanchantin2021general}
J.~Lanchantin, T.~Wang, V.~Ordonez, and Y.~Qi, ``General multi-label image
  classification with transformers,'' in \emph{Proceedings of the IEEE/CVF
  Conference on Computer Vision and Pattern Recognition}, 2021, pp.
  16\,478--16\,488.

\bibitem{yuan2021tokens}
L.~Yuan, Y.~Chen, T.~Wang, W.~Yu, Y.~Shi, Z.~Jiang, F.~E. Tay, J.~Feng, and
  S.~Yan, ``Tokens-to-token vit: Training vision transformers from scratch on
  imagenet,'' \emph{arXiv preprint arXiv:2101.11986}, 2021.

\bibitem{jin2016annotation}
J.~Jin and H.~Nakayama, ``Annotation order matters: Recurrent image annotator
  for arbitrary length image tagging,'' in \emph{International Conference on
  Pattern Recognition}.\hskip 1em plus 0.5em minus 0.4em\relax IEEE, 2016, pp.
  2452--2457.

\bibitem{chen2019multi}
Z.-M. Chen, X.-S. Wei, P.~Wang, and Y.~Guo, ``Multi-label image recognition
  with graph convolutional networks,'' in \emph{Proceedings of the IEEE
  Conference on Computer Vision and Pattern Recognition}, 2019, pp. 5177--5186.

\bibitem{zhang2017mixup}
H.~Zhang, M.~Cisse, Y.~N. Dauphin, and D.~Lopez-Paz, ``mixup: Beyond empirical
  risk minimization,'' in \emph{International Conference on Learning
  Representations}, 2018.

\bibitem{thulasidasan2019mixup}
S.~Thulasidasan, G.~Chennupati, J.~Bilmes, T.~Bhattacharya, and S.~Michalak,
  ``On mixup training: Improved calibration and predictive uncertainty for deep
  neural networks,'' in \emph{Advances in Neural Information Processing
  Systems}, 2019.

\bibitem{zhang2019bag}
Z.~Zhang, T.~He, H.~Zhang, Z.~Zhang, J.~Xie, and M.~Li, ``Bag of freebies for
  training object detection neural networks,'' \emph{arXiv preprint
  arXiv:1902.04103}, 2019.

\bibitem{islam2020feature}
M.~A. Islam, M.~Kowal, K.~G. Derpanis, and N.~D. Bruce, ``Feature binding with
  category-dependant mixup for semantic segmentation and adversarial
  robustness,'' in \emph{British Machine Vision Conference}, 2020.

\bibitem{verma2019manifold}
V.~Verma, A.~Lamb, C.~Beckham, A.~Najafi, I.~Mitliagkas, D.~Lopez-Paz, and
  Y.~Bengio, ``Manifold mixup: Better representations by interpolating hidden
  states,'' in \emph{International Conference on Machine Learning}.\hskip 1em
  plus 0.5em minus 0.4em\relax PMLR, 2019, pp. 6438--6447.

\bibitem{wang2019baseline}
Q.~Wang, N.~Jia, and T.~P. Breckon, ``A baseline for multi-label image
  classification using an ensemble of deep convolutional neural networks,'' in
  \emph{2019 IEEE International Conference on Image Processing (ICIP)}.\hskip
  1em plus 0.5em minus 0.4em\relax IEEE, 2019, pp. 644--648.

\bibitem{lin2014microsoft}
T.-Y. Lin, M.~Maire, S.~Belongie, J.~Hays, P.~Perona, D.~Ramanan,
  P.~Doll{\'a}r, and C.~L. Zitnick, ``Microsoft coco: Common objects in
  context,'' in \emph{European Conference on Computer Vision}.\hskip 1em plus
  0.5em minus 0.4em\relax Springer, 2014, pp. 740--755.

\bibitem{chua2009nus}
T.-S. Chua, J.~Tang, R.~Hong, H.~Li, Z.~Luo, and Y.~Zheng, ``Nus-wide: a
  real-world web image database from national university of singapore,'' in
  \emph{Proceedings of the ACM International Conference on Image and Video
  Retrieval}, 2009, pp. 1--9.

\bibitem{russakovsky2015imagenet}
O.~Russakovsky, J.~Deng, H.~Su, J.~Krause, S.~Satheesh, S.~Ma, Z.~Huang,
  A.~Karpathy, A.~Khosla, M.~Bernstein \emph{et~al.}, ``Imagenet large scale
  visual recognition challenge,'' \emph{International Journal of Computer
  Vision}, vol. 115, no.~3, pp. 211--252, 2015.

\bibitem{chen2018order}
S.-F. Chen, Y.-C. Chen, C.-K. Yeh, and Y.-C.~F. Wang, ``Order-free rnn with
  visual attention for multi-label classification,'' in \emph{AAAI Conference
  on Artificial Intelligence}, 2018.

\bibitem{li2018attentive}
L.~Li, S.~Wang, S.~Jiang, and Q.~Huang, ``Attentive recurrent neural network
  for weak-supervised multi-label image classification,'' in \emph{2018 ACM
  Multimedia Conference on Multimedia Conference}.\hskip 1em plus 0.5em minus
  0.4em\relax ACM, 2018, pp. 1092--1100.

\bibitem{zhu2017learning}
F.~Zhu, H.~Li, W.~Ouyang, N.~Yu, and X.~Wang, ``Learning spatial regularization
  with image-level supervisions for multi-label image classification,'' in
  \emph{Proceedings of the IEEE/CVF Conference on Computer Vision and Pattern
  Recognition}, 2017, pp. 5513--5522.

\bibitem{chen2020disentangling}
Z.~Chen, Q.~Cui, X.-S. Wei, X.~Jin, and Y.~Guo, ``Disentangling, embedding and
  ranking label cues for multi-label image recognition,'' \emph{IEEE
  Transactions on Multimedia}, 2020.

\bibitem{johnson2015love}
J.~Johnson, L.~Ballan, and L.~Fei-Fei, ``Love thy neighbors: Image annotation
  by exploiting image metadata,'' in \emph{International Conference on Computer
  Vision}, 2015, pp. 4624--4632.

\bibitem{li2017improving}
Y.~Li, Y.~Song, and J.~Luo, ``Improving pairwise ranking for multi-label image
  classification,'' in \emph{Proceedings of the IEEE/CVF Conference on Computer
  Vision and Pattern Recognition}, 2017, pp. 3617--3625.

\end{thebibliography}
\end{document}